\documentclass[sigconf]{acmart}
\AtBeginDocument{%
  }

\usepackage{xcolor}   
\usepackage{makecell} 
\usepackage{multirow} 
\usepackage{graphicx}
\usepackage{caption}
\usepackage{tabularx}

\usepackage[table,xcdraw]{xcolor}
\usepackage{siunitx}
\newcolumntype{Y}{>{\raggedright\arraybackslash}X}

\setcopyright{acmlicensed}

\copyrightyear{2026}
\acmYear{2026}
\setcopyright{cc}
\setcctype{by}
\acmConference[MM '26]{Proceedings of the 34th ACM International Conference on Multimedia}{November 10--14, 2026}{Rio de Janeiro, Brazil}
\acmBooktitle{Proceedings of the 34th ACM International Conference on Multimedia (MM '26), November 10--14, 2026, Rio de Janeiro, Brazil}
\acmDOI{10.1145/3767308.3835303}
\acmISBN{979-8-4007-2213-4/2026/11}

\begin{document}


\title{Beyond Uniform Restoration: Empowering All-in-One Restoration with Pixel-Level Multimodal Guidance}


\author{Chunxiao Liu}
\orcid{0009-0007-0275-7567}
\affiliation{%
  \institution{Xiaomi Corporation}
  \city{Beijing}
  \country{China}
}
\email{liuchunxiao@xiaomi.com}

\author{Wei Liu}
\orcid{0009-0003-6935-2231}
\affiliation{%
  \institution{Xiaomi Corporation}
  \city{Beijing}
  \country{China}
}
\email{liuwei67@xiaomi.com}

\author{Anbin Xiong}
\orcid{0009-0009-7411-0003}
\affiliation{%
  \institution{Xiaomi Corporation}
  \city{Beijing}
  \country{China}
}
\email{xionganbin@xiaomi.com}

\author{Erli Meng}
\orcid{0009-0009-3748-126X}
\affiliation{%
  \institution{Xiaomi Corporation}
  \city{Beijing}
  \country{China}
}
\email{mengerli@xiaomi.com}

\begin{abstract}
All-in-one image restoration is a unified low-level vision task that aims to effectively recover high-quality images from inputs degraded by various types and levels of corruption using a single model. Recent works have achieved remarkable progress by learning degradation-adaptive prompts or network architectures. However, these methods typically apply a uniform restoration strategy across the entire image, neglecting the fact that different regions may suffer from distinct degradation types and varying degrees of severity. In contrast, we propose to perform restoration at the pixel level, thereby enabling more fine-grained and precise control over the restoration process. Specifically, we present MGN-AIR, a novel pixel-level restoration framework for all-in-one image restoration. Our approach first learns to estimate a pixel-level visual prompt. Then, it leverages both textual and visual prompts to provide global and local degradation cues, guiding the model on where to look and how to restore at each pixel. We conduct extensive experiments on multiple all-in-one image restoration benchmarks, covering a wide range of tasks including denoising, deraining, deblurring, dehazing, desnowing, and low-light enhancement. Experimental results demonstrate that our proposed method consistently and significantly outperforms existing approaches.
\end{abstract}






\maketitle

\section{Introduction}
\label{sec:intro}
As a fundamental low-level vision task, image restoration focuses on recovering high-quality images from low-quality images through removing rain, haze, noise, etc. Due to the co‑existence of diverse degradations in real‑world images, learning a unified all‑in‑one restoration model to cope with diverse degradation types has drawn growing attention and achieved considerable advances.


\begin{figure}[tbp]
\centering
\includegraphics[width=0.5\textwidth]{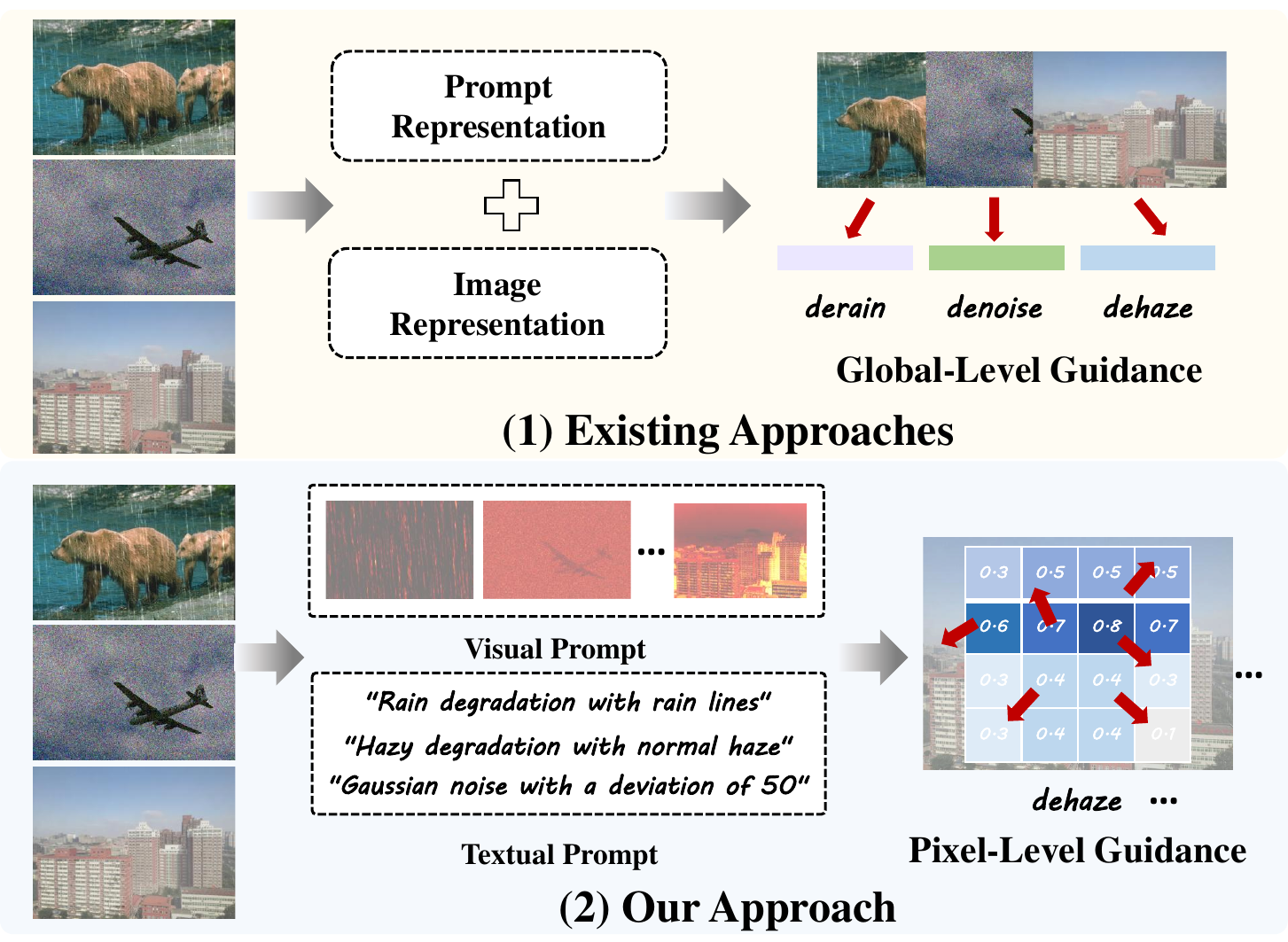}
\caption{Existing methods typically integrate prompt to provide global-level guidance for image restoration. In contrast, our approach enables fine-grained, pixel-level restoration by leveraging both textual and visual prompts that encodes global degradation and intensity of local degradation.}
\label{intro}
\vspace{-0.5cm}
\end{figure}


Existing all-in-one image restoration works can be roughly categorized into prompt-based approaches and adaptive network based approaches. Prompt-based approaches mainly focus on incorporating task priors to guide the degradation process. Textual prompts \cite{conde2024instructir,guo2024onerestore,yan2025textual}, visual prompts \cite{potlapalli2023promptir,luo2025visual} and multimodal prompts \cite{luo2024controlling, liao2025prompt} have been developed in image restoration networks and thus greatly improved the performance. These prompts are generally encoded by pretrained large models \cite{radford2021learning} or learnable parameters, guiding the restoration network to learn task-specific representations. Another branch of image restoration methods focuses on designing networks adaptive to various image restoration tasks, which aligns the feature space with a unified parameter space. For example, some works \cite{yang2024all,yang2024language,zamfir2025complexity,yu2024multi} introduce mixture-of-experts (MoE) architectures to learn a task-adaptive routing strategy, assigning degradations to the most suitable experts. External tasks like degradation classification \cite{huuniversal} and contrastive learning \cite{guo2024onerestore} are also employed to discriminate representations from different restorations. 

However, degradation patterns exhibit non-uniform distribution due to distinct degradation types and intensity. As shown in Fig.\ref{intro}, most regions in noisy images are severely corrupted, while only local regions in rainy images are rain-affected. This phenomenon indicates that image restoration requires more fine-grained guidance. For rainy images, restoration efforts should focus primarily on the rain-affected areas, while preserving the content of unaffected regions. Conversely, all regions in hazy images require restoration, but the degree of correction varies, some areas need only minor adjustments, while others demand more extensive correction. Existing methodologies typically treat the entire degraded image uniformly, neglecting the intrinsic variability in degradation and hence leading to suboptimal restoration outcomes.

To address the above issue, we propose a Multimodal Guidance Network for all-in-one image restoration, denoted as MGN-AIR, which performs more fine-grained and precise controls over the restoration process, thereby achieving more effective image restoration. Specifically, MGN-AIR first employs a visual prompt generation module to learn a visual prompt that reflects the local intensity of degradation. Next, in the multimodal guidance module, the learned visual prompt is incorporated with textual prompt to jointly predict a restoration matrix, which tells the model where to look and how to restore at each pixel. Lastly, the restoration matrix guides the image restoration in pixel-level restoration module, where heavily degraded regions primarily leverage contextual information from neighboring areas, while less-degraded regions rely more on recurring structural patterns to further enhance restoration quality. The main contributions are summarized as follows:
\begin{itemize}
    \item We propose a method to precisely perform image restoration at the pixel level. Compared with global-level restoration strategies, our design delivers more localized guidance, thereby enabling better adaptability to diverse degradation types and varying intensities.
    \item We design a multimodal guidance network for pixel-level all-in-one image restoration, consisting of a Visual Prompt Generation Module (VPGM), a Multimodal Guidance Module (MGM) and a Pixel-Level Restoration Module (PLRM), to apply pixel-level image restoration through predicting local visual prompt, generating pixel-level restoration guidance and executing pixel-level restoration.  
    \item Extensive experiments conducted on all-in-one image restoration benchmarks show that our network achieves state-of-the-art performance in all-in-one image restoration, especially for composite degradation benchmark, 1.51 dB PSNR improvement is obtained in comparison with recent works. 
\end{itemize}

\section{Related Work}
\label{sec:rw}
\noindent \textbf{Single Task Image Restoration.}
Restoring the high-quality clean image from its low-quality degraded version is an ill-posed issue. Early works primarily focus on designing task-specific networks based on hand-crafted priors \cite{lehtinen2018noise2noise,liang2021swinir,tai2017memnet,wang2022uformer,zamir2022restormer,zhang2017learning}. Recently, deep learning-based approaches apply convolutional \cite{zhang2019residual,zhang2017learning,tai2017memnet,chen2022simple} or Transformer \cite{chen2022cross,liu2021swin,wang2022uformer,zamir2022restormer} architectures, achieving remarkable progress on various image restoration tasks, such as deraining \cite{chen2021robust,li2019heavy,li2018recurrent,lin2024dual,wang2019erl}, dehazing \cite{chen2020pmhld,das2020fast,liu2019griddehazenet,qin2020ffa,shao2020domain,wu2021contrastive}, deblurring \cite{chen2024unsupervised,cho2021rethinking,kupyn2018deblurgan,kupyn2019deblurgan,park2020multi,zhang2020deblurring}, denoising \cite{huang2022casapunet,lin2023unsupervised,pan2022real,pan2023random,ren2021adaptive,ren2022enhanced} and low-light enhancement \cite{guo2020zero,ma2022toward,wu2022uretinex}. The convolutional networks aggregate local neighboring information to restore the degraded regions. In contrast, transformer-based networks capture global dependencies via self-attention mechanism, in which similar patterns play a more important role in restoration. The typical work is Restormer \cite{zamir2022restormer} that leverages transformer architectures to enhance image restoration tasks in a lightweight manner. However, these works are primarily designed for single-degradation, limiting their generalization capability in all-in-one image restoration.

\begin{figure*}[htbp]
\centering
\includegraphics[width=0.9\textwidth]{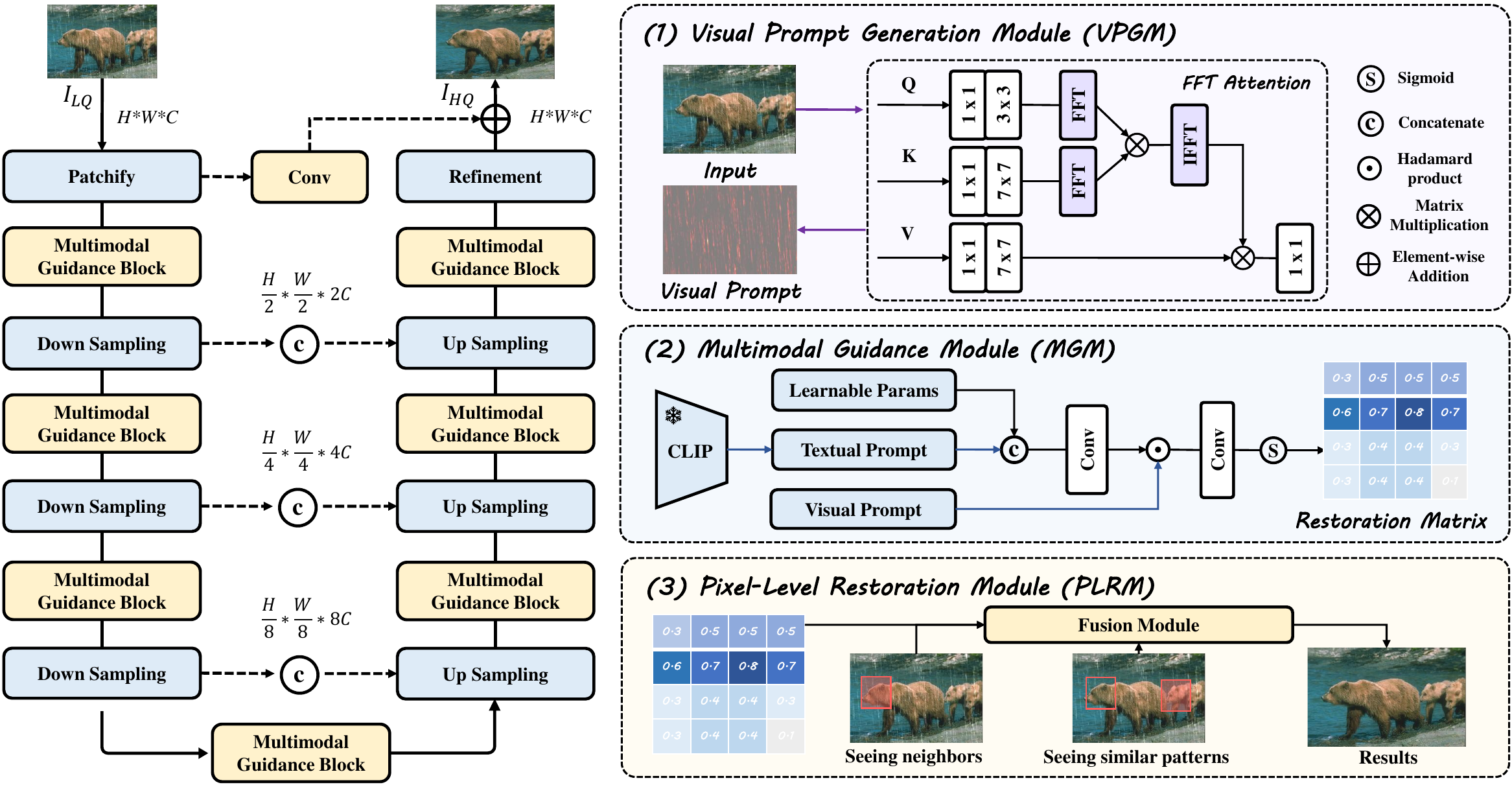}
\caption{Overall framework of our MGN-AIR. The left part is the U-net architecture, which consists of our proposed multimodal guidance block (MGB). The right part describes the detail of MGB. (1) Generating visual prompt supervised by the difference between degradad image and clean image; (2) Estimating pixel-wise restoration matrix guided by multimodal prompts; (3) performing pixel-level restoration based on restoration matrix.}
\label{method}
\vspace{-0.3cm}
\end{figure*}

\noindent \textbf{All-In-One Image Restoration.}
The typical all-in-one image restoration framework is encoder-decoder architecture. Based on this structure, various prompts have been proposed to enhance performance, including textual prompt, visual prompt and multimodal prompt. These prompts serve as an adaptive guidance module in the restoration process. For textual prompting, most works utilize the pretrained models as guidance to better understand task-specific context. InstructIR \cite{conde2024instructir} is the first approach that leverages real human-written instructions to guide the overall restoration process. For visual prompting, a representative work is PromptIR \cite{potlapalli2023promptir}, which enhances Restormer \cite{zamir2022restormer} by learning prompt representation and integrating it into the U-Net architecture \cite{ronneberger2015u}. For multimodal prompting, DA-CLIP \cite{luo2024controlling} designs a controller that firstly employs pretrained CLIP as text and image encoder, and then integrates the prompt embedding into image restoration models via cross attention. CyclicPrompt \cite{liao2025prompt} incorporates weather knowledge and conditional vectors into the visual prompt, jointly learning context-aware representations with textual prompt. Apart from prompt-based approaches, a large number of works focus on designing unified networks using adaptive learning strategies, such as multi-task collaboration \cite{yang2024all,yang2024language,zamfir2025complexity,yu2024multi}, self-contrastive learning \cite{li2022all,wu2025beyond}, adding task uncertainty regularization \cite{wu2025debiased}, considering degradation-aware feature perturbation \cite{tian2025degradation}, and applying different operations for hard and easy patches \cite{jiang2025cat}, etc.

Unlike prior approaches designed to adapt to diverse degradation types, our work pioneers a fine‑grained pixel‑wise network jointly driven by degradation type and intensity, making it well‑suited for realistic, complex, and non‑uniform degradations.

\section{Method}
\label{sec:method}

In this work, we present MGN-AIR, a multimodal guidance network for all-in-one image restoration. As illustrated in Fig.\ref{method}, the overall framework is a classic encoder-decoder network, in which each level is comprised of our proposed Multimodal Guidance Block (MGB), which firstly generates the visual prompt, and then integrates textual prompt and lastly performs pixel-level restoration.

\subsection{Overview}
Our framework builds upon a U-net like encoder-decoder architecture \cite{potlapalli2023promptir,zamir2022restormer,tian2025degradation,chen2022simple}. Given a degraded image $I_{LQ}\in \mathbb{R}^{H \times W \times 3}$ as input, our framework aims to restore a high-quality clean image $I_{HQ} \in \mathbb{R}^{H \times W \times 3}$. Here, $H \times W$ denotes the height and width of the image. The encoder‑decoder network comprises four levels, each equipped with a MGB. The encoder network takes a high-resolution image as input, the spatial resolution gradually decreases, while the channel increases through pixel shuffle. At the decoder stage, these changes are reversed to progressively restore the original resolution via pixel unshuffle. Encoder features at each level are fused with the decoder features at the same resolution via skip connections. 

Concretely, at the $i$-th level, the MGB takes features $F^{i-1}$ from previous level as input, performing pixel-level restoration spatially. The MGB mainly comprised of three modules. The first module is visual prompt generation module (VPGM) that generates pixel-level visual prompt $V^i$ in a learnable manner. The next module is multimodal guidance module (MGM) that learns to integrate both visual and textual prompt to provide multimodal guidance $P^i$ in a fine-grained granularity. At last, pixel-level restoration module (PLRM) executes distinct restoration at each pixel guided by $P^i$, producing intermediate image feature $F^i_{s}$. The overall process are: 

\begin{equation}
\begin{aligned}
 V^i & = \mathrm{VPGM}(F^{i-1}); \\
 P^i & = \mathrm{MGM}(T, V^i); \\
 F^i_{s} & = \mathrm{PLRM}(P^i, F^{i-1}). \\
\end{aligned}
\end{equation} 

After that, MGB takes intermediate features as input and applies a sequence of Transformer Blocks. In this block, We sequentially perform Channel Attention and feed‑forward network (FFN) to feature $F^i_{s}$, adding the output back to the original input feature via residual connection, producing latent feature $F^{i}$. The overall structure of MGB enables us to gradually and adaptively recover degraded region guided by pixel-level information. Note that in the final decoder layer, we directly adopt the visual prompt of the preceding layer instead of applying VPGM.

\begin{equation}
\begin{aligned}
    \hat{F}^{i} &= \text{ChannelAttn}\left(F^i_{s}\right) + F^i_{s}, \\
    F^{i} &= \text{FFN}\left(\hat{F}^{i}\right) + \hat{F}^{i}.
\end{aligned}
\end{equation}

\subsection{Visual Prompt Generation Module (VPGM)}
The goal of VPGM is to estimate the visual prompt that indicates the degradation map of each pixel. Taking the degraded image $\mathbf{I}_{LQ} \in \mathbb{R}^{H \times W \times C}$ as input, where $C$ denotes the channel number. For the first level, the channel number equals to 3, while other levels equals to the feature dimension. The VPGM firstly employs a $1 \times 1$ convolution, and then applies a window-based self-attention mechanism to generate visual prompt based on spatial information. To reduce the computation cost, the matrix multiplication between queries $Q'$ and keys $K'$ is performed in the Fourier domain \cite{kong2023efficient}. Subsequently, the computed features are then transformed back into the spatial domain, performing maxtrix multiplication with values $V'$ after Softmax:
\begin{equation}
\begin{aligned}
Q' &= x W_Q, \quad K' = x W_K, \quad V' = x W_V;  \\
\hat{Q} &= \mathcal{F}(Q'), \quad \hat{K} = \mathcal{F}(K');  \\
A &= \text{Softmax}\left( \frac{1}{\sqrt{d}} \mathcal{F}^{-1}\left( \hat{Q} \cdot \hat{K} \right)\right);  \\
V &= A \cdot V'.
\end{aligned}
\end{equation}

\noindent where, $x \in \mathbb{R}^{r \times r \times C}$ denotes the content within each window, the window size $r$ is set as 32. $Q', K', V' \in \mathbb{R}^{r \times r \times d}$ represent query, key and value in self-attention mechanism, and \( W_Q, W_K, W_V \in \mathbb{R}^{C \times d} \) are corresponding learnable weights. $\mathcal{F}$ and $\mathcal{F}^{-1}$ represent FFT and inverse FFT, respectively. Subsequently, the $1 \times 1$ convolution and SiLU activation function are applied, followed by Sigmoid function to predict the degradation map $V \in \mathbb{R}^{H \times W}$ of each pixel, which is our proposed visual prompt.

Notably, the visual prompt is generated explicitly only at the first level, while prompts at subsequent levels are learned implicitly. Explicit learning at the initial stage requires computing the pixel-wise difference between the degraded image and its ground-truth as supervision. This difference directly reflects the spatial intensity of degradation, thereby providing precise, pixel-level guidance for restoration. In contrast, deeper levels benefit from implicit learning, which allows the model to adaptively capture higher-order dependencies without relying on direct supervision. Specifically, we first compute the spatial pixel-wise difference and then average it across the channel dimension. After applying min-max normalization, we obtain a degradation map, which serves as the ground truth pixel-level visual prompt for the first level, formulated as follows:
\begin{equation}
\begin{aligned}
X & = \frac{1}{C}\sum_{i=1}^{C}\left\| I^{i}_{HQ} - I^{i}_{LQ} \right\|_1; \\
\hat{X} & = \frac{X - X_{\min}}{X_{\max} - X_{\min}}.
\end{aligned}
\end{equation}

\noindent where, $i$ denotes the channel index, $I^{i}_{HQ}$ and $I^{i}_{LQ}$ denote the $i$-th channel in ground truth image and degraded image, respectively.

\subsection{Multimodal Guidance Module (MGM)}
Recent works \cite{conde2024instructir,guo2024onerestore,yan2025textual} have demonstrated the potential of textual prompts in all-in-one image restoration as it facilitates discriminating different types of degradation. However, in real-world scenarios, degradation exhibits diverse types and intensities, which demands more fine-grained, pixel-level control instead of the global guidance provided by existing textual prompts. Consequently, providing a more fine-grained guidance is crucial for this field. 

To address the above issue, we propose a novel module MGM that leverages both visual and textual prompts for pixel-level image restoration. Specifically, the textual prompt conveys high-level semantic information about the degradation type (e.g., “raindrops”), thereby determining the distribution or pattern of degradation across the image. Complementarily, the visual prompt provides spatially precise guidance by indicating the severity of degradation at each pixel. For instance, for a rainy image, the textual prompt specifies that the corruption is caused by raindrops, while the visual prompt reveals where and how severely each pixel is affected.

Specifically, we employ the frozen CLIP \cite{radford2021learning} text encoder to extract a textual representation $T\in \mathbb{R}^{1 \times d}$, where $d=512$ is the embedding dimension of CLIP. The visual prompt $V \in \mathbb{R}^{H \times W}$ is estimated by the preceding module VPGM. We first aggregate textual information by applying an average pooling on the textual prompt, producing a textual feature as global guidance. Then, a learnable scaling parameter $\beta$ is concatenated with the global guidance, followed by a 1D convolution. To adapt this global feature to the spatial domain, we replicate the textual feature to match the spatial resolution $H \times W$, yielding $T^p \in \mathbb{R}^{H \times W \times C}$, where $C=32$. Subsequently, the modulated textual prompt and the visual prompt are combined via element-wise multiplication, followed by a $1 \times 1$ convolution and Sigmoid activation. The resulting restoration matrix $P \in [0,1]^{H \times W \times C}$ employs both local and global guidance to jointly determine the restoration at each pixel. The overall process is formulated as:
\begin{equation}
\begin{aligned}
T^p & =\mathrm{Concat}[\mathrm{Avg}(T);\beta]; \\
V^p & = \mathrm{Conv1D}(T^p) \odot V ; \\
P & = \mathrm{Sigmoid}(\mathrm{Conv2D}(V^p)).
\end{aligned}
\end{equation}

\begin{table*}[htbp]
\centering
\footnotesize 
\setlength{\tabcolsep}{0.03pt}  
\caption{\textit{Comparison state-of-the-art on composited degradations dataset CDD11 \cite{guo2024onerestore}.} PSNR (dB, ↑) and  \colorbox[RGB]{238,244,251}{SSIM (↑)} are reported on the full RGB images. Our method consistently outperforms existing models on both single and composited degradations.}
\begin{tabularx}{\textwidth}{@{}l
>{\centering\arraybackslash}X
>{\centering\arraybackslash\cellcolor[RGB]{238,244,251}}X 
>{\centering\arraybackslash}X 
>{\centering\arraybackslash\cellcolor[RGB]{238,244,251}}X 
>{\centering\arraybackslash}X 
>{\centering\arraybackslash\cellcolor[RGB]{238,244,251}}X 
>{\centering\arraybackslash}X 
>{\centering\arraybackslash\cellcolor[RGB]{238,244,251}}X 
>{\centering\arraybackslash}X 
>{\centering\arraybackslash\cellcolor[RGB]{238,244,251}}X
>{\centering\arraybackslash}X 
>{\centering\arraybackslash\cellcolor[RGB]{238,244,251}}X
>{\centering\arraybackslash}X
>{\centering\arraybackslash\cellcolor[RGB]{238,244,251}}X 
>{\centering\arraybackslash}X 
>{\centering\arraybackslash\cellcolor[RGB]{238,244,251}}X 
>{\centering\arraybackslash}X 
>{\centering\arraybackslash\cellcolor[RGB]{238,244,251}}X 
>{\centering\arraybackslash}X 
>{\centering\arraybackslash\cellcolor[RGB]{238,244,251}}X 
>{\centering\arraybackslash}X 
>{\centering\arraybackslash\cellcolor[RGB]{238,244,251}}X
>{\centering\arraybackslash}X 
>{\centering\arraybackslash\cellcolor[RGB]{238,244,251}}X}
\toprule
\multirow{2}{*}{Method} & \multicolumn{8}{c}{CDD11-Single} & \multicolumn{10}{c}{CDD11-Double} & \multicolumn{4}{c}{CDD11-Triple} & \multicolumn{2}{c}{\multirow{2}{*}{Average}}  \\ \cline{2-23}
 & \multicolumn{2}{c}{Low(L)} & \multicolumn{2}{c}{Haze(H)} & \multicolumn{2}{c}{Rain(R)} & \multicolumn{2}{c}{Snow(S)} & \multicolumn{2}{c}{L+H} & \multicolumn{2}{c}{L+R} & \multicolumn{2}{c}{L+S} & \multicolumn{2}{c}{H+R} & \multicolumn{2}{c}{H+S} & \multicolumn{2}{c}{L+H+R} & \multicolumn{2}{c}{L+H+S} & \multicolumn{2}{c}{} \\
\midrule
AirNet \cite{li2022all} & 24.83 & .778 & 24.21 & .951 & 26.55 & .891 & 26.79 & .919 & 23.23 & .779 & 22.82 & .710 & 23.29 & .723 & 22.21 & .868 & 23.29 & .901 & 21.80 & .708 & 22.24 & .725 & 23.75 & .814 \\
PromptIR \cite{potlapalli2023promptir} & 26.32 & .805 & 26.10 & .969 & 31.56 & .946 & 31.53 & .960 & 24.49 & .789 & 25.05 & .771 & 24.51 & .761 & 24.54 & .924 & 23.70 & .925 & 23.74 & .752 & 23.33 & .747 & 25.90 & .850 \\
WGWSNet \cite{zhu2023learning} & 24.39 & .774 & 27.90 & .982 & 33.15 & .964 & 34.43 & .973 & 24.27 & .800 & 25.06 & .772 & 24.60 & .765 & 27.23 & .955 & 27.65 & .960 & 23.90 & .772 & 23.97 & .771 & 26.96 & .863 \\
WeatherDiff \cite{ozdenizci2023restoring} & 23.58 & .763 & 21.99 & .904 & 24.85 & .885 & 24.80 & .888 & 21.83 & .756 & 22.69 & .730 & 22.12 & .707 & 21.25 & .868 & 21.99 & .868 & 21.23 & .716 & 21.04 & .698 & 22.49 & .799 \\
OneRestore \cite{guo2024onerestore} & 26.48 & \textbf{.826} & 32.52 & .990 & 33.40 & .964 & 34.31 & .973 & 25.79 & .822 & 25.58 & .799 & 25.19 & .789 & 29.99 & .957 & 30.21 & .964 & 24.78 & .788 & 24.90 & .791 & 28.47 & .878 \\
MoCE-IR \cite{zamfir2025complexity} & 27.26 & .824 & 32.66 & .990 & 34.31 & .970 & 35.91 & .980 & 26.24 & .817 & 26.25 & .800 & 26.04 & .793 & 29.93 & .964 & 30.19 & .970 & 25.41 & .789 & 25.39 & .790 & 29.05 & .881 \\
MGN-AIR  & \textbf{27.30} & .824 & \textbf{36.06} & \textbf{.994} & \textbf{36.53} & \textbf{.981} & \textbf{38.40} & \textbf{.986} & \textbf{27.01} & \textbf{.826} & \textbf{26.81} & \textbf{.808} & \textbf{26.06} & \textbf{.802} & \textbf{32.54} & \textbf{.978} & \textbf{33.01} & \textbf{.980} & \textbf{26.08} & \textbf{.801} & \textbf{26.33} & \textbf{.802} & \textbf{30.56} & \textbf{.889} \\
\bottomrule
\end{tabularx}
\label{mixed-degradation}
\end{table*}

\subsection{Pixel-Level Restoration Module (PLRM)}
In the preceding module, a pixel-level restoration matrix is estimated by fusing visual and textual prompts. This module, referred to as PLRM, is designed to perform image restoration adaptively. Specifically, different restoration strategies are applied depending on local degradation severity. 
For instance, in mildly degraded pixels, such as those affected by haze or low light, effective restoration benefits from both local contextual information and recurring structural patterns in the image. In contrast, for heavily degraded pixels, such as those corrupted by rain or noise, only neighboring pixels remain reliable, and hence require a more localized approach.

To address the above issue, we propose to utilize similar patterns and neighbor information based on restoration matrix. For heavily degraded pixels, we combine standard $3 \times 3$ 2D convolutions with dilated convolutions to enhance receptive field, in case of degradation span large spatial areas. Inspired by the principle of the attention mechanism, the query feature to be restored heavily relies on similar patterns in the image. We employ a series of spatial attention blocks to address less-degraded pixels, operating within an $8 \times 8$ local window to reduce computational cost. The combination of two strategies ensures effective contextual modeling even under severe corruption.
\begin{equation}
\begin{aligned}
\hat{F} & = W_{c} \cdot \mathrm{Conv2D}(F) + W_{d} \cdot \mathrm{DilaConv}(F); \\
F_{s} & = P \odot \hat{F} + (1-P) \odot \mathrm{SpatialAttn}(F). \\
\end{aligned}
\end{equation}
where, $W_{c}$ and $W_{d}$ denote learnable parameters for $\mathrm{Conv2D}$ and $\mathrm{DilaConv}$ operation, $P$ denotes the restoration matrix learned by previous module MGM.

\subsection{Optimization}
Our approach takes into account both perceptual quality and pixel-level fidelity of image restoration, and hence employs pixel-wise loss $\mathcal{L}_{L_{1}}$, perceptual loss $\mathcal{L}_{Percep}$ and SSIM loss $\mathcal{L}_{SSIM}$ as the objective function. 
Here, $L_{1}$ loss measures absolute pixel differences. Additionally, we employ the $L_{1}$ loss between estimated visual prompt and the real degradation map as an auxiliary loss to supervise the visual prompt learning, denoted as $\mathcal{L}_{Aux}$. The SSIM loss accounts for structural patterns and local dependencies, which is more aligned with human visual perception. This loss is a differentiable approximation of the negative SSIM metric, which quantifies the perceived similarity between two images by comparing their luminance, contrast, and structural information. The perceptual loss measures the discrepancy between high-level feature representations of the predicted and ground-truth images. The representations are generally extracted from pre-trained VGG network \cite{simonyan2014very}, capturing semantic and textural similarities that better reflect human judgment. Therefore, the overall loss function is:
\begin{equation}
\mathcal{L}_{Total} = \mathcal{L}_{L_{1}} + \mathcal{L}_{SSIM} + \mathcal{L}_{Percep} + \mathcal{L}_{Aux}.
\end{equation}

\section{Experiment}
\label{sec:exp}
\begin{table*}[h]
\centering
\footnotesize 
\setlength{\tabcolsep}{0.5pt}  
\caption{\textit{Comparison to state-of-the-art on five degradations.} PSNR (dB, ↑) and  \colorbox[RGB]{238,244,251}{SSIM (↑)} are reported on the full RGB images. Our method achieves a new state-of-the-art on average across all benchmarks. The best performances are highlighted.}
\begin{tabularx}{\textwidth}{@{}l c
>{\centering\arraybackslash}X
>{\centering\arraybackslash\cellcolor[RGB]{238,244,251}}X 
>{\centering\arraybackslash}X 
>{\centering\arraybackslash\cellcolor[RGB]{238,244,251}}X 
>{\centering\arraybackslash}X 
>{\centering\arraybackslash\cellcolor[RGB]{238,244,251}}X 
>{\centering\arraybackslash}X 
>{\centering\arraybackslash\cellcolor[RGB]{238,244,251}}X 
>{\centering\arraybackslash}X 
>{\centering\arraybackslash\cellcolor[RGB]{238,244,251}}X
>{\centering\arraybackslash}X 
>{\centering\arraybackslash\cellcolor[RGB]{238,244,251}}X 
@{}}
\toprule
\multirow{2}{*}{Method} & \multirow{2}{*}{Venue} & \multicolumn{2}{c}{Dehazing} & \multicolumn{2}{c}{Deraining} & \multicolumn{2}{c}{Denoising} & \multicolumn{2}{c}{Deblurring} & \multicolumn{2}{c}{Low-Light} & \multicolumn{2}{c}{\multirow{2}{*}{Average}} \\ \cline{3-12}
 &  & \multicolumn{2}{c}{SOTS} & \multicolumn{2}{c}{Rain100L} & \multicolumn{2}{c}{BSD68$_{\sigma=25}$} & \multicolumn{2}{c}{GoPro} & \multicolumn{2}{c}{LOLv1} & \multicolumn{2}{c}{} \\ \midrule
NAFNet \cite{chen2022simple} & ECCV'22 & 25.23 & .939 & 35.56 & .967 & 31.02 & .883 & 26.53 & .808 & 20.49 & .809 & 27.76 & .881 \\
DGUNet \cite{mou2022deep} & CVPR'22 & 24.78 & .940 & 36.62 & .971 & 31.10 & .883 & 27.25 & .837 & 21.87 & .823 & 28.32 & .891 \\
SwinIR \cite{liang2021swinir} & ICCV'21 & 21.50 & .891 & 30.78 & .923 & 30.59 & .868 & 24.52 & .773 & 17.81 & .723 & 25.04 & .835 \\
Restormer \cite{zamir2022restormer} & CVPR'22 & 24.09 & .927 & 34.81 & .962 & 31.49 & .884 & 27.22 & .829 & 20.41 & .806 & 27.60 & .881 \\
MambaIR \cite{guo2024mambair} & ECCV'24 & 25.81 & .944 & 36.55 & .971 & 31.41 & .884 & 28.61 & .875 & 22.49 & .832 & 28.97 & .901 \\ 
DL \cite{fan2019general} & TPAMI'19 & 20.54 & .826 & 21.96 & .762 & 23.09 & .745 & 19.86 & .672 & 19.83 & .712 & 21.05 & .743 \\
Transweather \cite{valanarasu2022transweather} & CVPR'22 & 21.32 & .885 & 29.43 & .905 & 29.00 & .841 & 25.12 & .757 & 21.21 & .792 & 25.22 & .836 \\
TAPE \cite{liu2022tape} & ECCV'22 & 22.16 & .861 & 29.67 & .904 & 30.18 & .855 & 24.47 & .763 & 18.97 & .621 & 25.09 & .801 \\
AirNet \cite{li2022all} & CVPR'22 & 21.04 & .884 & 32.98 & .951 & 30.91 & .882 & 24.35 & .781 & 18.18 & .735 & 25.49 & .847 \\
IDR \cite{zhang2023ingredient} & CVPR'23 & 25.24 & .943 & 35.63 & .965 & \textbf{31.60} & .887 & 27.87 & .846 & 21.34 & .826 & 28.34 & .893 \\
PromptIR \cite{potlapalli2023promptir} & NeurIPS'23 & 26.54 & .949 & 36.37 & .970 & 31.47 & .886 & 28.71 & .881 & 22.68 & .832 & 29.15 & .904 \\
InstructIR \cite{conde2024instructir} & ECCV'24 & 27.10 & .956 & 36.84 & .973 &  31.40 & .887 & 29.40 & .886 & 23.00 & .836 & 29.55 & .908 \\
AdaIR \cite{cui2025adair} & ICLR'25 & 30.53 & .978 & 38.02 & .981 & 31.35 & .889 & 28.12 & .858 & 23.00 & .845 & 30.20 & .910 \\
MoCE-IR \cite{zamfir2025complexity} & CVPR'25 & 30.48 & .974 & 38.04 & .982 & 31.34 & .887 & 30.05 & .899 & 23.00 & .852 & 30.58 & .919 \\ 
VLU-Net \cite{zeng2025vision} & CVPR'25 & 30.84 & .980 & 38.54 & .982 & 31.43 & \textbf{.891} & 27.46 & .840 & 22.29 & .833 & 30.11 & .905 
\\ 
DFPIR \cite{tian2025degradation} & CVPR'25 & 31.64 & .979 & 37.62 & .978 & 31.29 & .889 & 28.82 & .873 & \textbf{23.82} & .843 & 30.64 & .913 \\
MGN-AIR (Ours) & —— & \textbf{31.74} & \textbf{.982} & \textbf{39.47} & \textbf{.986} & 31.12 & .884 & \textbf{31.18} & \textbf{.917} & 23.67 & \textbf{.862} & \textbf{31.44} & \textbf{.926} \\
\bottomrule
\end{tabularx}
\label{five-degradation}
\end{table*}

In this work, we closely follow previous all-in-one image restoration works \cite{conde2024instructir,duan2024uniprocessor,potlapalli2023promptir,jiang2025cat,li2022all,zamir2022restormer,guo2024onerestore}, evaluating our proposed method on various experimental settings, including mixed degradations and typical all-in-one settings. In these settings, a unified restoration model is trained to address multiple degradation types. The evaluation metrics are PSNR and SSIM, which are standard and widely used for assessing image quality in prior works. Methods with best performance are highlighted in bold.

\subsection{Experimental Settings}
\textbf{Datasets.}
For mixed degradation setting, we employ CDD11 dataset \cite{guo2024onerestore} as our benchmark to evaluate the effectiveness on composite degraded images. For typical all-in-one restoration setting, the dataset is composed of multiple task-specific datasets. The BSD400 \cite{arbelaez2010contour} and WED \cite{ma2016waterloo} datasets are combined as the training dataset for image denoising, adding Gaussian noise at different levels $\sigma \in [15, 20, 50]$ to create noisy images. The SOTS \cite{li2018benchmarking} dataset is used for image dehazing. Rain100L \cite{yang2020learning} is adopted for image deraining. The GoPro \cite{nah2017deep} and LOLv1 \cite{weideep} datasets are employed for deblurring and low-light enhancement, respectively. Evaluation is performed on standard benchmark BSD68 \cite{martin2001database} for image denoising, and corresponding test sets for other tasks. A single unified model is trained on the union of all the aforementioned training datasets and evaluated directly across all restoration tasks without task-specific adaptation.

\noindent \textbf{Implementation Details.} 
We propose a unified framework MGN-AIR for all-in-one image restoration. Following the architectural design of PromptIR \cite{potlapalli2023promptir} and DFPIR \cite{tian2025degradation}, our approach adopts a 4-level encoder-decoder structure, with the number of transformer blocks at each level set to [4, 6, 6, 8] from level-1 to level-4, respectively. All experiments are implemented by PyTorch and conducted on a single NVIDIA H200 GPU. In mixed degradation setting, the model is trained for 300 epochs with an initial learning rate of $2\times 10^{-4}$. In typical all-in-one setting, the model is trained for 80 epochs. During the training phase, we randomly crop the full degraded image into $128 \times 128$ patches, and perform the data augmentation using random horizontal and vertical flips. The batch size is set as 25. The network is optimized by Adam optimizer, with parameters $\beta_{1}$ equal to 0.9 and $\beta_{2}$ equal to 0.999.

\subsection{Experimental Results}
\noindent \textbf{All-in-One: Mixed-degradation.}
To better simulate realistic degradation scenarios, recent studies \cite{guo2024onerestore,zamfir2025complexity} have trained all-in-one restoration models on more challenging images, where multiple degradation types co-occur within a single image. The typical benchmark is CDD11, in which rain, haze, snow, and low illumination four degradation types are regarded as base degradations, and their combinations are also included as composite degradation scenarios, yielding a total of eleven unique restoration tasks. In this benchmark, the degradation type and degradation level of each pixel might be different. The results in Tab.\ref{mixed-degradation} show that our approach MGN-AIR surpasses recent work MoCE-IR \cite{tian2025degradation} and OneRestore \cite{guo2024onerestore} on average PSNR by 1.51 dB and 2.09 dB, respectively, validating the effectiveness of our approach in handling more complex and realistic scenarios. In contrast, existing methods typically adopt a global restoration strategy, which struggles to achieve fine-grained, pixel-adaptive image restoration.

\begin{table}[]
\centering
\footnotesize 
\setlength{\tabcolsep}{0.1pt}  
\caption{\textit{Comparison to state-of-the-art on three degradations.} PSNR (dB, ↑) and \colorbox[RGB]{238,244,251}{SSIM (↑)} are reported on the full RGB images. Our method achieves a new state-of-the-art on average across all benchmarks.}
\begin{tabularx}{0.48\textwidth}{@{}l
>{\centering\arraybackslash}X
>{\centering\arraybackslash\cellcolor[RGB]{238,244,251}}X 
>{\centering\arraybackslash}X 
>{\centering\arraybackslash\cellcolor[RGB]{238,244,251}}X 
>{\centering\arraybackslash}X 
>{\centering\arraybackslash\cellcolor[RGB]{238,244,251}}X 
>{\centering\arraybackslash}X 
>{\centering\arraybackslash\cellcolor[RGB]{238,244,251}}X 
@{}}
\toprule
\multirow{2}{*}{Method} & \multicolumn{2}{c}{Dehazing} & \multicolumn{2}{c}{Deraining} & \multicolumn{2}{c}{Denoising} & \multicolumn{2}{c}{\multirow{2}{*}{Average}} \\ \cline{2-7}
 &   \multicolumn{2}{c}{SOTS} & \multicolumn{2}{c}{Rain100L} & \multicolumn{2}{c}{BSD68$_{15-50}$}  & \multicolumn{2}{c}{} \\ \midrule
LPNet \cite{gao2019dynamic} & 20.84 & .828 & 24.88 & .784 & 24.17 & .693 & 23.30 & .768 \\
FDGAN \cite{dong2020fd} & 24.71 & .929 & 29.89 & .933 & 28.50 & .851 & 27.70 & .904 \\
DL \cite{fan2019general} & 26.92 & .931 & 32.62 & .931 & 30.12 & .838 & 29.89 & .900 \\
MPRNet \cite{zamir2021multi} & 25.28 & .955 & 33.57 & .954 & 30.66 & .862 & 29.84 & .924 \\
AirNet \cite{li2022all} & 27.94 & .962 & 34.90 & .967 & 31.06 & .873 & 31.30 & .934 \\
Restormer \cite{zamir2022restormer} & 30.43 & .975 & 36.55 & .975 & 30.97 & .869 & 32.65 & .940 \\
PromptIR \cite{potlapalli2023promptir} & 30.58 & .974 & 36.37 & .972 & 31.12 & .873 & 32.69 & .940 \\
InstructIR \cite{conde2024instructir} & 30.22 & .959 & 37.98 & .978 & 31.32 & .876 & 33.17 & .938 \\
AdaIR \cite{cui2025adair} & 31.06 & .980 & 38.64 & .983 & 31.25 & .876 & 33.65 & .946 \\
MoCE-IR \cite{zamfir2025complexity} & 31.34 & .979 & 38.57 & .984 & 31.25 & .873 & 33.72 & .945 \\
DFPIR \cite{tian2025degradation} & 31.87 & .980 & 38.65 & .982 & \textbf{31.29} & \textbf{.878} & 33.94 & .947 \\
MGN-AIR  & \textbf{33.21} & \textbf{.987} & \textbf{39.62} & \textbf{.986} & 31.00 & .868 &  \textbf{34.61} & \textbf{.947} \\ \bottomrule
\end{tabularx}
\label{three-degradation}
\end{table}

\noindent \textbf{All-in-One: Five Degradations.} 
We compare our unified image restoration approach with classic works, including image-only models like AirNet \cite{li2022all}, PromptIR \cite{potlapalli2023promptir}, multimodal models like DA-CLIP \cite{luo2024controlling}, InstructIR \cite{conde2024instructir}, and recent works like multi-expert network MoCE-IR \cite{zamfir2025complexity}, VLM-based network VLU-Net \cite{zeng2025vision}, and perturbation-aware network DFPIR \cite{tian2025degradation}. All these models are trained on a comprehensive dataset of five degradation types, including denoising, dehazing, deraining, deblurring and low-light enhancement. The results in Tab.\ref{five-degradation} demonstrate that our proposed network MGN-AIR, achieving the best performance on average PSNR and SSIM, with 2.29 dB improvement in PSNR over baseline work PromptIR \cite{potlapalli2023promptir}. Compared with most recent work DFPIR \cite{tian2025degradation}, our approach significantly improves the overall performance, especially on deblurring and deraining tasks, achieving 2.36 dB and 1.85 dB improvement in PSNR, which validates the effectiveness of our approach. Notably, compared with multi-expert model MoCE-IR \cite{zamfir2025complexity}, our approach greatly outperforms it using only one expert, which demonstrates the effectiveness of fine-grained guidance.

\begin{table}[htbp]
\centering
\caption{\textit{Ablation study on the multimodal guidance block design.} PSNR (dB, ↑) and \colorbox[RGB]{238,244,251}{SSIM (↑)} are reported.}
\footnotesize 
\setlength{\tabcolsep}{0.22pt}  
\begin{tabularx}{0.48\textwidth}{@{}l
>{\centering\arraybackslash}X
>{\centering\arraybackslash\cellcolor[RGB]{238,244,251}}X 
>{\centering\arraybackslash}X 
>{\centering\arraybackslash\cellcolor[RGB]{238,244,251}}X 
>{\centering\arraybackslash}X 
>{\centering\arraybackslash\cellcolor[RGB]{238,244,251}}X 
>{\centering\arraybackslash}X 
>{\centering\arraybackslash\cellcolor[RGB]{238,244,251}}X 
@{}}
\toprule
\multirow{2}{*}{Method} & \multicolumn{2}{c}{Dehazing} & \multicolumn{2}{c}{Deraining} & \multicolumn{2}{c}{Denoising} & \multicolumn{2}{c}{\multirow{2}{*}{Average}} \\ \cline{2-7}
 &   \multicolumn{2}{c}{SOTS} & \multicolumn{2}{c}{Rain100L} & \multicolumn{2}{c}{BSD68$_{15-50}$}  & \multicolumn{2}{c}{} \\ \midrule
w/o MGB  & 31.23 & .980 & 38.17 & .982 & \textbf{31.23} & \textbf{.877} & 33.54 & .946 \\
w/o visual prompt & 30.89 & .978 & 37.34 & .980 & 30.83 & .865 & 33.02 & .941 \\
w/o textual prompt  & 31.31 & .980 & 37.37 & .980 & 30.80 & .865 & 33.16 & .942 \\
w/o multi-layer block & 31.91 & .982 & 38.84 & .984 & 30.89 & .867 & 33.88 & .944 \\
MGN-AIR & \textbf{33.21} & \textbf{.987} & \textbf{39.62} & \textbf{.986} & 31.00 & .868 & \textbf{34.61} & \textbf{.947} \\
\bottomrule
\end{tabularx}
\label{ab:component}
\vspace{-0.1cm}
\end{table}

\begin{figure*}[htbp]
\centering
\includegraphics[width=0.9\textwidth]{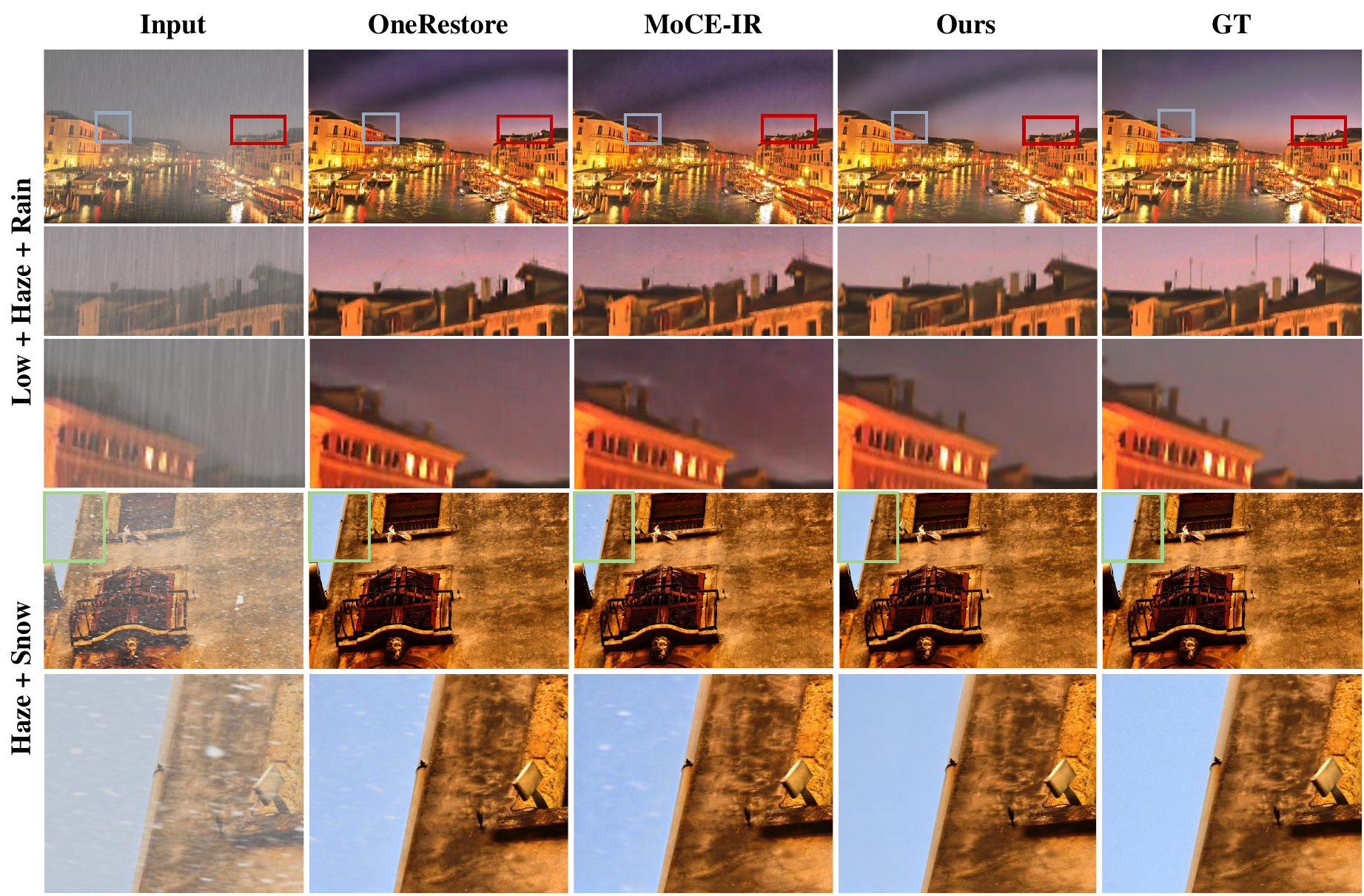}
\caption{Visual results of MoCE-IR \cite{tian2025degradation}, OneRestore \cite{guo2024onerestore} and our approach on CDD11 dataset \cite{guo2024onerestore}. We provide the global image and local detail grounded by different colors for better visualization. MGN-AIR effectively removes haze, rain and snow streaks while preserving local details in the image.}
\label{vis}
\vspace{-0.5cm}
\end{figure*}

\noindent \textbf{All-in-One: Three Degradations.} 
We also evaluate the overall average performance across three image restoration tasks: denoising at three noise levels ($\sigma$ = 15, 25, 50), dehazing, and deraining. For denoising, results on BSD68 are averaged over the three noise levels and reported as BSD68$_{15-50}$. Experimental results in Tab.\ref{three-degradation} show that our proposed approach MGN-AIR can significantly improve the performance of dehazing and deraining. The average PSNR of our approach also outperforms existing methods. The performance on denoising task is slightly lower than state-of-the-arts, which is primarily due to severe disturbance of both local and global information by noise.


\noindent \textbf{Effectiveness of each component.}
We conduct extensive ablation experiments to validate the effectiveness of our proposed multimodal guidance block, reporting the results of training an all-in-one model on combined datasets across three restoration tasks. The experiment settings contain the following five variants: (1) w/o MGB: Only using U-net structure, removing multimodal guidance block; (2) w/o visual prompt: Removing the VPGM, and removing visual prompt in MGM; (3) w/o textual prompt: Removing textual prompt in MGM; (4) w/o multi-layer block:  Only using multimodal guidance block at the first layer, removing this block from subsequent layers; and (5) MGN-AIR: Using all the proposed module. As shown in Tab.\ref{ab:component}, removing any component in MGN-AIR will greatly influence the performance. Concretely, removing visual prompt will largely reduce the effect, which demonstrates the fine-grained pixel-level guidance is crucial for image restoration. Moreover, removing textual prompt will also influence performance, since different degradation types indicate different distributions. Compared with w/o MGB setting, only preserving MGB at the first layer can yield a 0.34 dB PSNR improvement, and adding multiple MGB will further achieve 0.73 dB PSNR imporvement, which validates the effectiveness of our proposed block. 


\noindent \textbf{Complexity comparisons:} We compare the computational complexity of our method against standard baselines, including PromptIR \cite{potlapalli2023promptir} and AirNet \cite{li2022all}. To demonstrate that our performance gains stem from the efficacy of the proposed MGB block rather than merely an increase in model capacity, we designed experiments focusing on two aspects: (a) \textit{Capacity reduction}: We compress the number of layers within the MGB to create a smaller variant, denoted as \textit{Ours-s}; and (b) \textit{Structural substitution}: We replace our MGB block with alternative simple operations, such as spatial attention or gated networks, while maintaining a comparable parameter count. As shown in Tab.~\ref{cost}, our proposed module consistently outperforms these variants, and our lightweight version \textit{Ours-s} outperforms baselines while keeps similar parameter counts, confirming that the improvements are attributed to the specific design of the MGB.

\begin{table}[H]
\vspace{-0.3cm}
\caption{\textit{Computation cost comparison.}}
\label{cost}
\vspace{-0.2cm}
\begin{tabular}{c c c c cc}
\hline
\setlength{\tabcolsep}{0.01pt}  
Model & \begin{tabular}[c]{@{}c@{}}Params\\ (M)\end{tabular} & \begin{tabular}[c]{@{}c@{}}FLOPs\\ (G)\end{tabular} & \begin{tabular}[c]{@{}c@{}}Inference \\ Latency\\ (ms)\end{tabular} & \multicolumn{2}{c}{Average} \\ \hline
PromptIR \cite{potlapalli2023promptir} & 32.97 & 121.08 & 31.09 & \multicolumn{1}{c}{32.69} & .940 \\ 
AirNet \cite{li2022all} & 5.77 & 230.66 & 28.83 & \multicolumn{1}{c}{31.30} & .934 \\
Spatial & 34.29 & 217.14 & 64.84 & \multicolumn{1}{c}{33.69} & .946 \\ 
Gated & 33.63 & 208.56 & 61.75 & \multicolumn{1}{c}{29.43} & .904 \\ 
Ours & 34.33 & 215.19 & 63.25 & \multicolumn{1}{c}{\textbf{34.61}} & \textbf{.947} \\ 
Ours-s & 26.61 & 133.42 & 35.84 & \multicolumn{1}{c}{\textbf{34.13}} & \textbf{.946} \\ \hline
\end{tabular}
\end{table}

\noindent \textbf{Visualization.}
To intuitively demonstrate the effectiveness of our method, we provide visualization results under the mixed degradation settings in Fig.\ref{vis}. In complex scenarios composited by haze, rain and low illumination, both MoCE-IR \cite{tian2025degradation} and OneRestore \cite{guo2024onerestore} perform image restoration at global level, resulting in noticeable errors in local regions. In contrast, our approach greatly preserves the local details while removing degradations. 
For example, in challenging scenes composited by haze, rain and low illumination, the antenna on the right rooftop, a subtle yet important structural detail, is largely neglected in prior works. Our method recovers it with high fidelity, whereas MoCE-IR exhibits considerable blurriness. Furthermore, on the left rooftop, existing methods exhibit over-smoothing, leading to white artifacts on the roof surface, while our approach effectively addresses this issue by pixel-level restoration. 
In certain scenes composited by haze and snow, our approach generates clear and sharp denoised outputs, while existing methods fail to remove snow that appears in the top-left corner of the image.

\section{Conclusion}
\label{sec:conlusion}
This paper presents MGN-AIR, a multimodal guidance network for fine-grained all-in-one image restoration, which operates the restoration process at the pixel level for adaptive restoration control. By jointly leveraging the learned visual and textual prompts, our method tailors the restoration operation for each pixel based on local intensity and global degradation type. Extensive experiments in diverse degradation settings show that MGN-AIR can significantly outperform existing methods, especially on scenarios with multiple composited degradations.

\bibliographystyle{ACM-Reference-Format}
\bibliography{sample-base}


\end{document}